\documentclass[a4paper, 10pt, conference]{ieeeconf}      
\usepackage{FG2019}

\usepackage{booktabs}
\usepackage{adjustbox}
\usepackage{cite}
\usepackage{subfigure}
\usepackage{comment}
\FGfinalcopy 

\IEEEoverridecommandlockouts                              
\def\FGPaperID{163978} 

\title{\LARGE \bf
Characterization of the Handwriting Skills \\as a Biomarker for Parkinson's Disease
}

\author{\parbox{16cm}{\centering
    {\large R. Castrill\'on$^{1,2}$, A. Acien$^3$, J.R. Orozco-Arroyave$^2$, A. Morales$^3$, J.F. Vargas$^2$, R. Vera-Rodr\'iguez$^3$, J. Fierrez$^3$, J. Ortega-Garcia$^3$, A. Villegas$^2$}\\
    {\normalsize
    $^1$Universidad Cat\'olica de Oriente, Rionegro, Colombia.
    $^2$Universidad de Antioquia, Medell\'in, Colombia\\
    $^3$Universidad Aut\'onoma de Madrid, Madrid, Espa\~na}}
    \thanks{This work was financed by CODI from University of Antioquia grant PRG2015-7683 and the Neurometrics project funded by UAM-Banco Santander.}
}

\begin{document}
\IEEEoverridecommandlockouts\pubid{\makebox[\columnwidth]{978-1-7281-0089-0/19/\$31.00~\copyright{}2019 IEEE \hfill}
\hspace{\columnsep}\makebox[\columnwidth]{ }}
\ifFGfinal
\thispagestyle{empty}
\pagestyle{empty}
\else
\author{Anonymous FG 2019 submission\\ Paper ID \FGPaperID \\}
\pagestyle{plain}
\fi

\maketitle

\begin{abstract}
In this paper we evaluate the suitability of handwriting patterns as potential biomarkers
to model Parkinson's disease (PD).  Although the study of PD is attracting the interest of many researchers around the world, databases to evaluate handwriting patterns are scarce and knowledge about 
patterns associated to PD is limited and biased to the existing datasets. This paper introduces a database with a total of 935 handwriting tasks collected from 55 PD patients and 94 healthy controls (45 young and 49 old). Three feature sets are extracted from the signals: neuromotor, kinematic, and 
non-linear dynamic. Different classifiers are used to discriminate between PD and healthy subjects: support vector machines, k-nearest neighbors, and a multilayer perceptron. The proposed features and classifiers enable to detect PD with accuracies between 81\% and 97\%. Additionally, new insights are presented on the utility of the studied features for monitoring and detecting PD.
\end{abstract}
\section{INTRODUCTION}
Parkinson's disease (PD) is a neurodegenerative disorder that occurs due to loss 
of dopamine, a neurotransmitter that helps in regulating muscle movements. 
The disease is chronic and progressive, and affects multiple areas of 
the central nervous system. 
PD is characterized by alterations of the motor system such as bradykinesia, 
resting tremor, muscular rigidity, and posture~\cite{jankovic2008parkinson}. 
There is no known cure for PD~\cite{conolly2014pharma} and its early diagnosis is crucial for more effective treatments~\cite{Becker2002early}.
The process to diagnose and evaluate the progression of PD is purely 
subjective~\cite{Rosenblum2013}. To evaluate the disease progression neurologists administer different clinical tests such as the MDS-UPDRS and H\&Y scales to the patients. 
These scales are usually limited to evaluate upper limb motor skills~\cite{Smits2014}. 
Also, small changes in the progression are not detectable through those scales. 
Besides, the probability of incorrect diagnosis based on such scales is around 25\% \cite{Drotar2016}. 
A more accurate assessment of motor activities would allow medical doctors and researchers 
to measure the disease progression and make timely decisions about the therapy. 

Handwriting analysis offers the possibility to assess and monitor those motor skills of 
PD patients. 
Different abnormal behaviors in handwriting are observed in PD patients.
For instance, micrographia occurs in 5\% of the patients before other 
motor symptoms appear, and about 30\% of the handwriting worsening cases are reported 
after the medical diagnosis~\cite{Thomas2017}. 
Handwriting tasks have significant advantages: they are simple, less intrusive, 
natural, do not need specialized infrastructure and can be administered remotely. 
There are several studies in the literature which are focused on the automatic
evaluation of handwriting of PD patients considering different tasks and machine learning
algorithms. 
Drotar et al. use kinematic and pressure analysis to classify between PD patients and 
healthy subjects. Using a population of 37 PD and 38 Healthy Controls (HC), they report classification accuracies 
of up to 82\%. 
The studies are carried out considering different tasks including spirals, sentences and characters~\cite{Drotar2016}. 
Mucha et al. proposed a new approach named ``Fractional Derivative'' to improve 
the classification considering kinematic handwriting signals extracted from drawings of a 
spiral. 
They use a database of 30 PD/36 HC and report a classification accuracy of 72.39\%~\cite{mucha2018fractional}. 
Using repetitive cursive loops and kinematic features for evaluation, Haremans et al. 
found a correlation of $r = -0.40$ between the handwriting measurements and the 
medical scales considering a corpus with 30 PD/15 HC~\cite{heremans2016handwriting}. 
From a population of 24 PD/20 HC, Kotsavasiloglou et al. report classification results 
of 91\% using kinematic features and entropy analysis from drawings of horizontal lines~\cite{kotsa2017machine}. Finally, Taleb et al. report results of 96.87\% when 
classifying between PD patients and HC subjects. Seven tasks are considered from a corpus 
with 16 PD/16 HC~\cite{Taleb2017feature}. All these efforts and others have been summarized in recent surveys~\cite{impedovo2018dynamic}~\cite{stefano2018review}.\\
\hspace{1cm} To the best of our knowledge, this paper introduces the largest database for the 
analysis of online handwriting of PD patients and control users including both young healthy and elderly people with similar age and educational level with respect to the PD patients. As we will see later, both groups of controls are important in order to characterize neuromotor diseases against typical neuromotor degradation caused by the age. Additionally, we propose to 
evaluate different tasks (individually and merged) modeled considering several
feature sets including kinematics, nonlinear dynamics, and neuromotor. The automatic
discrimination between PD patients and HC subjects is evaluated considering three different
classifiers: k-nearest neighbors (KNN), support vector machines (SVM), 
and a multilayer perceptron (MLP).

\begin{table*}[h!]
\caption{Demographic and clinical data of patients and controls}\label{tab:demographic}
\centering
\resizebox{0.7\textwidth}{!}{
\begin{tabular}{lcccccc}  
\toprule
\multicolumn{1}{c}{} & \multicolumn{2}{c}{PD Patients} & \multicolumn{2}{c}{EHC (Old Controls)} &
\multicolumn{2}{c}{YHC (Young Controls)}\\
\cmidrule(r){2-7}
& Male & Female & Male & Female & Male & Female\\
\midrule
Number & \phantom{..}20 & 35 & 27 & 22 & 27 & 18\\
Age & \phantom{.....}64$\pm$10.3 &	\phantom{...}58$\pm$12.9 & \phantom{...}66$\pm$11.1& \phantom{...}59$\pm$10.6 & \phantom{...}25$\pm$4.93	& \phantom{...}23$\pm$3.9\\	
Range of Age & \phantom{..}41--80 &		\phantom{}29--83 & \hspace{0.5mm}49--85	&	\hspace{0.5mm}43--83	&	\hspace{0.5mm}17--42	&	\hspace{0.5mm}19--32\\			
UPDRS & \phantom{...}39$\pm$19 &33$\pm$22&$-$ &$-$ &$-$ &$-$ \\
H\&Y & \phantom{...}2.3$\pm$0.5 &	\phantom{}2.4$\pm$0.8 & -- &$-$ &$-$ &$-$ \\							
\bottomrule
\end{tabular}}
\end{table*}

\section{DATA}
\subsubsection{Subjects}

A total of 55 non-demented PD patients (35 female) with an average age of 60 years (standard deviation $SD=$11.9) were enrolled for this study. Two sets of healthy participants are also considered,
one is formed with 49 elderly people (22 female), namely EHC, who are
matched in age and education level with respect to the patients, and the 
other one includes 45 young healthy controls (18 female), namely YHC,
with ages between 17 and 42 years. We consider that it is important to include both groups of controls to differentiate between patterns associated to the PD disease and patterns associated to the natural degradation of the neuromotor abilities with the age. 
During the recording sessions all of the patients were under the effect of their medication
(i.e., ON-state). 
Further details about the participants are presented in Table~\ref{tab:demographic}.

\subsubsection{Tasks}

The participants were asked to complete 17 different handwriting tasks following a template. 
The first tasks consisted of writing the letters \emph{l} and \emph{m} in a continuous and 
long trace. Other tasks include the digits (0 to 9), the ID, name and signature of the participant, 
a free sentence, and the alphabet. 
The other nine tasks consist of geometrical figures including an Archimedean spiral, a circle 
with and without a template, a house, two concentric rectangles, a rhombus, a cube, 
and the Rey-Osterrieth complex figure. 

\subsubsection{Acquisition system}

The handwriting signals were recorded using a commercial tablet Wacom Cintiq (13HD Touch, 180\,Hz 
of sampling frequency), which captures different signals including x-position, y-position, 
pressure, in-air movement, azimuth and altitude of the pen over the tablet, and writing time. Although all of these signals were captured with the tablet, this paper only considers experiments with the first
three signals.

\section{METHODOLOGY}

\subsection{Feature Extraction}

Kinematic, non-linear dynamics and neuromotor features are extracted from the signals.
Since there are tasks composed by several strokes, the features are extracted 
globally, i.e., per task, and also per stroke within each task.
Strokes are segmented according to the pen-down and pen-up movements. 
A total of eleven statistical functionals are computed from the features: 
mean value, median, standard deviation, 1st percentile, 99th percentile, 
difference between the 99th and 1st percentiles, maximum, minimum, kurtosis and skewness. 
This procedure results in a 921-dimensional feature vector per task containing a total of 
452 kinematic features, 354 nonlinear dynamics features and 115 neuromotor features. Although we have not included all features proposed in the literature, we consider that this feature set is representative of the state-of-the-art~\cite{stefano2018review}.

\subsubsection{Kinematic features}
Global features refer to features extracted from the whole handwriting task. Mean velocity, max acceleration, distance between adjacent points or total duration are examples of these features. Global feature sets have been used to characterize handwriting signatures for many years with good performance~\cite{Fierrez2008signature} and, more recently, to characterize swipe patterns~\cite{Acien2018Agegroups}. But they have never been studied to characterize PD, so in this paper we will analyze if they are suitable for this purpose.  Many global feature sets have been proposed in the online handwriting recognition literature~\cite{Serwadda2013benchmark,Martinez2014GlobalFeatures,fierrezSwipe}. In this paper we use the extended set proposed in~\cite{Martinez2014GlobalFeatures} which comprises 100 of the best performing global features adapted for online signatures.

\subsubsection{Nonlinear Dynamics features}

These are computed with the aim of modeling stability, non-stationarity in muscular movements
that cannot be accurately modeled with classical approaches like those based on kinematics.
There are studies that show the relationship between motor activities of handwriting and
chaotic processes~\cite{longstaff1999nonlinear}. 
The nonlinear approach has been successfully applied to model other bio-signals like 
voice~\cite{Travieso2017} and gait~\cite{Perez2018}. The first step in the analysis of non-linear dynamics is the reconstruction of the phase space. 
It allows the study of the dynamic behavior of a time series. 
Periodic signals exhibit closed trajectories in the phase space while non periodic 
signals show irregular and chaotic patterns. 
In this paper, different nonlinear features are extracted from the reconstructed attractor 
such as correlation dimension, Lempel-Ziv complexity, largest Lyapunov exponent, 
Hurst exponent, empirical mode decomposition, and entropy.
Other non linear features typically extracted to model handwriting signals are considered including 
the Shannon entropy, 2nd and 3rd order Renyi entropy, and the signal-to-noise ratio 
calculated using the conventional energy definition and the Teager-Kaiser 
energy~\cite{drotar2015decision}.

\subsubsection{Neuromotor features}
The Sigma-Lognormal model first introduced by~\cite{Reilly2009neuromotor} decomposes the velocity profile of human handwriting into stroke velocity signals with lognormal shape. The Sigma-Lognormal theory states that lognormal functions describe the rapid changes in the velocity of writing movements produced by neuromotor signals~\cite{Duval2015sigmalognormal}. According to this, the velocity signal $v_i (t)$ of each of these strokes, $i$, can be described with lognormal shape. The velocity profile can be used as a marker of neuronological disorders. Healthy patients tend to show velocity signals with less number of lognormals and stable bandwidths while the PD patients velocity signals show a large number of lognormals (due to poor motor control ability) and variable bandwidths. Up to 28 features are calculated and used to depict the neuromotor ability of the user according to space-based and time-based features from the lognormal parameters (see~\cite{Fischer2008signature} for all details). The neuromotor feature set is computed averaging the parameters of all lognormals from the handwriting task to obtain a single value for task/feature.

\subsection{Classification and parameter optimization}
Three different classifiers are tested in this study, KNN,
SVM and MLP.
The KNN consists of assigning certain class to a test sample according to the
number of nearest neighbors ($K$) that such a sample has in the feature space.
The second classifier considered is a radial basis function SVM (RBF-SVM). 
In this case two meta-parameters (the margin parameter $C$ and the bandwidth of the Gaussian 
kernel $\gamma$) need to be optimized in the training process with the aim of finding the
optimal hyperplane that better separates PD patients and HC subjects.
The third classifier is a fully connected feed-forward neural network with  
an input layer, several hidden layers, and an output layer. 
A single neuron in the MLP is able to separate the input space into two 
subspaces by a hyperplane which is defined by the weights and the threshold. 
The MLP classifier uses the back-propagation algorithm for the adaptation of the weights \cite{gil2009diagnosing}. 

\subsubsection*{Meta-parameter optimization}

The meta-parameters of the three classifiers are optimized following a similar strategy. The process 
consists of a leave-one-out cross-validation strategy in a grid-search over a set with
different candidate values for the meta-parameters. The optimal values are found considering
only one of the 17 tasks that the participants performed during the recording process. 
The other 16 tasks are considered as the test set.
In the case of the KNN the number of neighbors is optimized in the set $K \in \{3,5,7,9,11,15\}$;
the meta-parameters of the RBF-SVM, $C$ and $\gamma$, are optimized also in a grid search
where $C\in \{0.001, 0.01, 0.1, 1, 10, 1000, 2000, 10000\}$ and similarly
$\gamma \in \{10^{-6},10^{-5},10^{-4},0.01, 0.1, 1, 10, 1000 \}$.
The number of layers in the MLP is also optimized in a grid search within the set $\{ 5,15,30 \}$.

\begin{table*}[t]
\centering
\caption{Classification results per task using the RBF-SVM}
\resizebox{\textwidth}{!}{
\begin{tabular}{lcccccccccccc}  
\toprule
\multicolumn{1}{c}{} &
\multicolumn{12}{c}{Experiments and feature sets} \\
\cmidrule(r){2-13}
 & \multicolumn{4}{c}{YHC vs PD} & \multicolumn{4}{c}{EHC vs PD} & \multicolumn{4}{c}{YHC vs EHC}\\
\cmidrule(r){2-5}\cmidrule(r){6-9}\cmidrule(r){10-13}
&Kinematic & Non linear & Neuromotor &All &Kinematic & Non linear & Neuromotor &All&Kinematic & Non linear & Neuromotor &All \\
\midrule
Alphabet       & 91.8\% & 87.6\% & ---- & 90.7\% & 73.3\% & 71.3\% & ---- & 71.3\% & 88.3\% & 74.5\% & ---- & 77.7\% \\
Circletemplate & 70.3\% & 73.0\% & 59.5\%          & 71.6\% & 55.7\% & 65.7\% & 55.7\%        & 65.7\% & 68.4\% & 73.7\% & 65.8\%        & 75.0\% \\
Cube           & 84.2\% & 72.3\% & 70.3\%          & 76.2\% & 68.6\% & 65.7\% & 61.0\%        & 68.6\% & 76.6\% & 69.1\% & 62.8\%        & 60.6\% \\
Freewriting    & 83.7\% & 80.6\% & ---- & 81.6\% & 65.7\% & 63.7\% & ---- & 68.6\% & 79.3\% & 75.0\% & ---- & 72.8\% \\
House          & 81.2\% & 77.2\% & 72.3\%          & 82.2\% & 76.2\% & 72.4\% & 65.7\%        & 72.4\% & 67.0\% & 60.6\% & 62.8\%        & 72.3\% \\
ID             & 78.8\% & 84.8\% & 69.7\%          & 83.8\% & 68.0\% & 66.0\% & 55.3\%        & 67.0\% & 86.2\% & 78.7\% & 73.4\%        & 75.5\% \\
Name           & 91.0\% & 81.0\% & 69.0\%          & 82.7\% & 63.5\% & 63.5\% & 62.5\%        & 60.6\% & 84.0\% & 76.6\% & 59.6\%        & 80.9\% \\
Numbers        & 83.8\% & 84.8\% & 68.7\%          & 72.0\% & 65.0\% & 61.2\% & 61.2\%        & 65.0\% & 73.4\% & 73.4\% & 64.9\%        & 75.5\% \\
Line1          & 86.7\% & 72.0\% & 77.3\%          & 90.0\% & 62.0\% & 62.0\% & 47.9\%        & 69.0\% & 78.9\% & 69.7\% & 73.7\%        & 76.3\% \\
Line2          & 76.0\% & 77.3\% & 54.7\%          & 85.9\% & 64.8\% & 56.3\% & 53.5\%        & 62.0\% & 80.3\% & 80.3\% & 71.1\%        & 80.3\% \\
Rectangles     & 83.2\% & 79.2\% & 68.3\%          & 71.3\% & 61.0\% & 65.7\% & 50.5\%        & 73.3\% & 62.8\% & 67.0\% & 61.7\%        & 63.8\% \\
Rey            & 87.5\% & 77.1\% & ---- & 90.6\% & 68.7\% & 64.6\% & ----& 73.7\% & 80.6\% & 77.4\% & ---- & 78.5\% \\
Rhombus        & 72.3\% & 78.2\% & 65.3\%          & 74.3\% & 54.3\% & 61.0\% & 54.3\%        & 59.0\% & 72.3\% & 64.9\% & 64.9\%        & 70.2\% \\
Signature      & 87.9\% & 81.8\% & 70.7\%          & 93.9\% & 71.8\% & 69.9\% & 65.0\%        & 75.7\% & 78.7\% & 62.8\% & 58.5\%        & 76.6\% \\
Spiral         & 71.3\% & 73.3\% & 68.3\%          & 77.2\% & 61.0\% & 61.9\% & 59.0\%        & 63.8\% & 58.5\% & 73.4\% & 55.3\%        & 61.7\% \\
Spiraltemplate & 78.2\% & 77.2\% & 79.2\%          & 80.2\% & 60.6\% & 62.5\% & 53.8\%        & 72.1\% & 69.9\% & 64.5\% & 63.4\%        & 66.7\% \\
Circle*         & 78.2\% & 77.2\% & 73.3\%          & 73.3\% & 61.9\% & 58.1\% & 55.2\%        & 67.6\% & 78.7\% & 66.0\% & 72.3\%& 75.5\%\\
\midrule
Opt. params. &C=1000 & C= 10 & C=1 & C=2000 & C=2000 & C=100 & C=1 & C=100 & C=10 & C=1 & C=2000& C=10\\
 &$\gamma=10^{-5}$ & $\gamma=0.01$ & $\gamma=10^{-5}$          & $\gamma=10^{-6}$ & $\gamma=10^{-6}$ & $\gamma=0.001$ & $\gamma=10^{-4}$& $\gamma=10^{-5}$ & $\gamma=0.001$ & $\gamma=0.01$ & $\gamma=10^{-5}$& $\gamma=0.001$\\
\bottomrule
\multicolumn{13}{l}{* This result corresponds to the training process for optimizing the meta-parameters.}
\end{tabular}}
\label{tab:svm5foldfull}
\end{table*}

\section{EXPERIMENTAL RESULTS}

Table \ref{tab:svm5foldfull} shows the best results obtained for the different classification experiments: 
YHC vs PD; EHC vs PD; and YHC vs EHC. All of them correspond to the RBF-SVM, and 
similar results were found with the MLP. The results obtained with the KNN were not
satisfactory. Note that similar results are obtained with the kinematic features and with the combination of all of the features. \emph{Alphabet} and \emph{Signature} tasks present the best classification accuracy: over 90\% for YHC vs PD and over 70\% for EHC vs PD. On the other hand, \emph{Circle with template} and \emph{Rectangles} present performances below 60\% in some cases. In the tasks of greatest complexity (\emph{Alphabet}, \emph{Freewriting} and \emph{Rey}) the neuromotor characteristics were not calculated because these features are designed for tasks of simple shapes and short strokes.
In addition to the tests on individual tasks, all of the tasks are combined to create a 
generalized model. This combination is performed in late-fusion strategy, i.e., the
scores of the classifier (RBF-SVM in all of the cases) are combined according to the mean rule to obtain a new score \cite{fierrezMCS}. The results are reported in Table~\ref{tab:resultscores}.  

\begin{table}[htb!]
\centering
\caption{Results of the classification combining all tasks}
\resizebox{0.48\textwidth}{!}{
\begin{tabular}{lcccc}
\toprule
Set of features&Class.& YHC vs PD & EHC vs PD & YHC vs EHC\\
\midrule
\midrule
Kinematics&SVM &96.9\%&78.3\%&94.4\%\\
&KNN &93.7\%&74.9\%&87.5\%\\
&MLP &96.9\%&73.4\%&94.4\%\\
\midrule
Non linear&SVM &95.6\%&78.3\%&91.6\%\\
&KNN &94.3\%&61.4\%&86.2\%\\
&MLP &96.9\%&78.3\%&94.4\%\\
\midrule
Neuromotor&SVM &88.0\%&51.6\%&81.9\%\\
&KNN &89.3\%&65.2\%&79.1\%\\
&MLP &81.8\%&59.9\%&81.9\%\\
\midrule
All& SVM&96.9\%&81.7\%&97.2\%\\
&KNN &96.9\%&73.4\%&87.5\%\\
&MLP &96.9\%&78.3\%&94.4\%\\
\bottomrule
\multicolumn{5}{l}{Note: the \emph{Circle} task was not considered, as it was used before for training.}
\end{tabular}}
\label{tab:resultscores}
\end{table}

The results obtained with the fusion strategy are better than those obtained with individual models for each task. Further research in this topic may help to clarify which are the most important tasks to discriminate between PD and healthy subjects, and what is the best way to combine them.

\begin{figure}[htb!]
\centering
\subfigure[Kinematics.]{\includegraphics[width=0.23\textwidth]{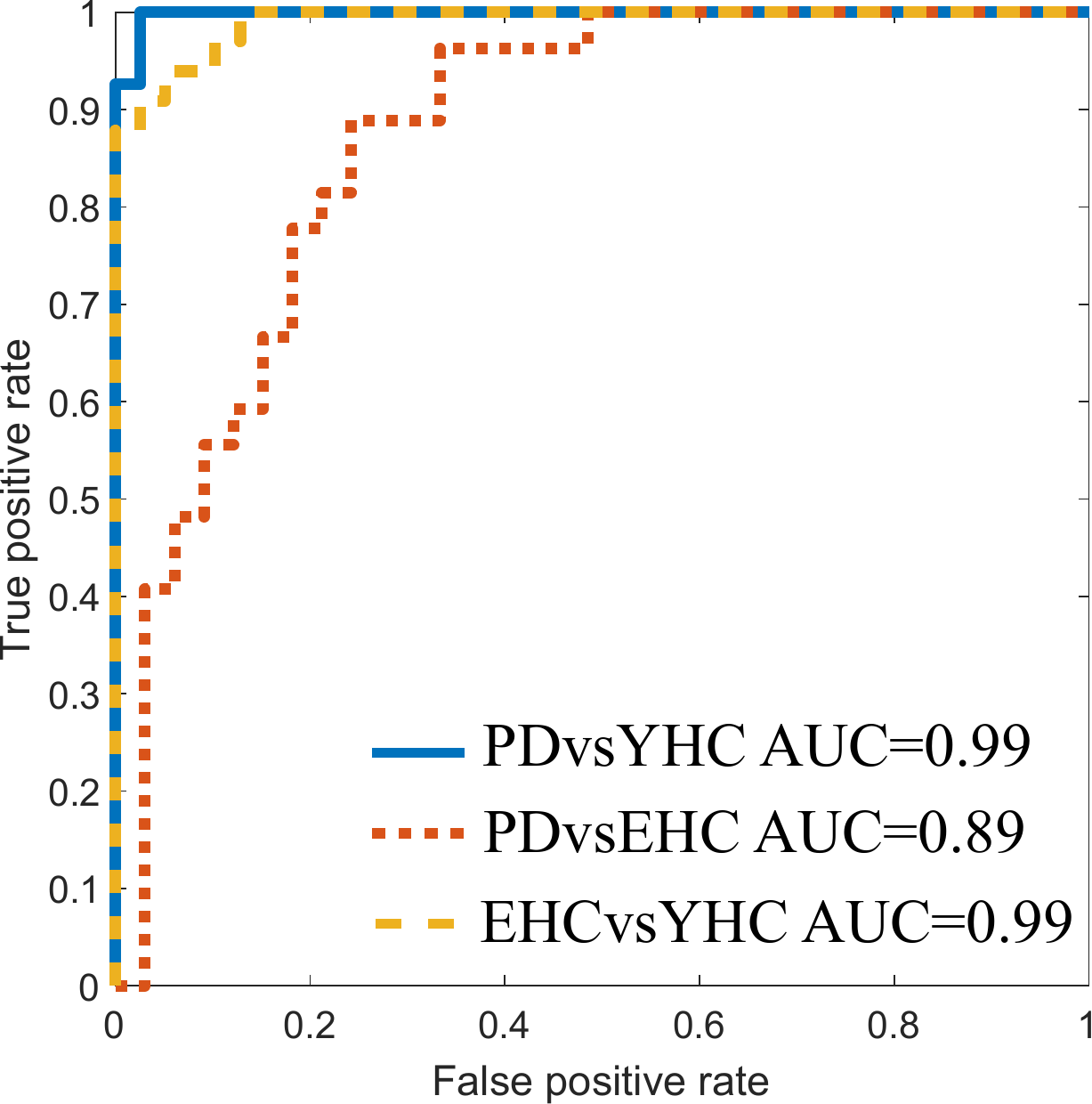}}
\subfigure[Dynamics Non Linear.]{\includegraphics[width=0.23\textwidth]{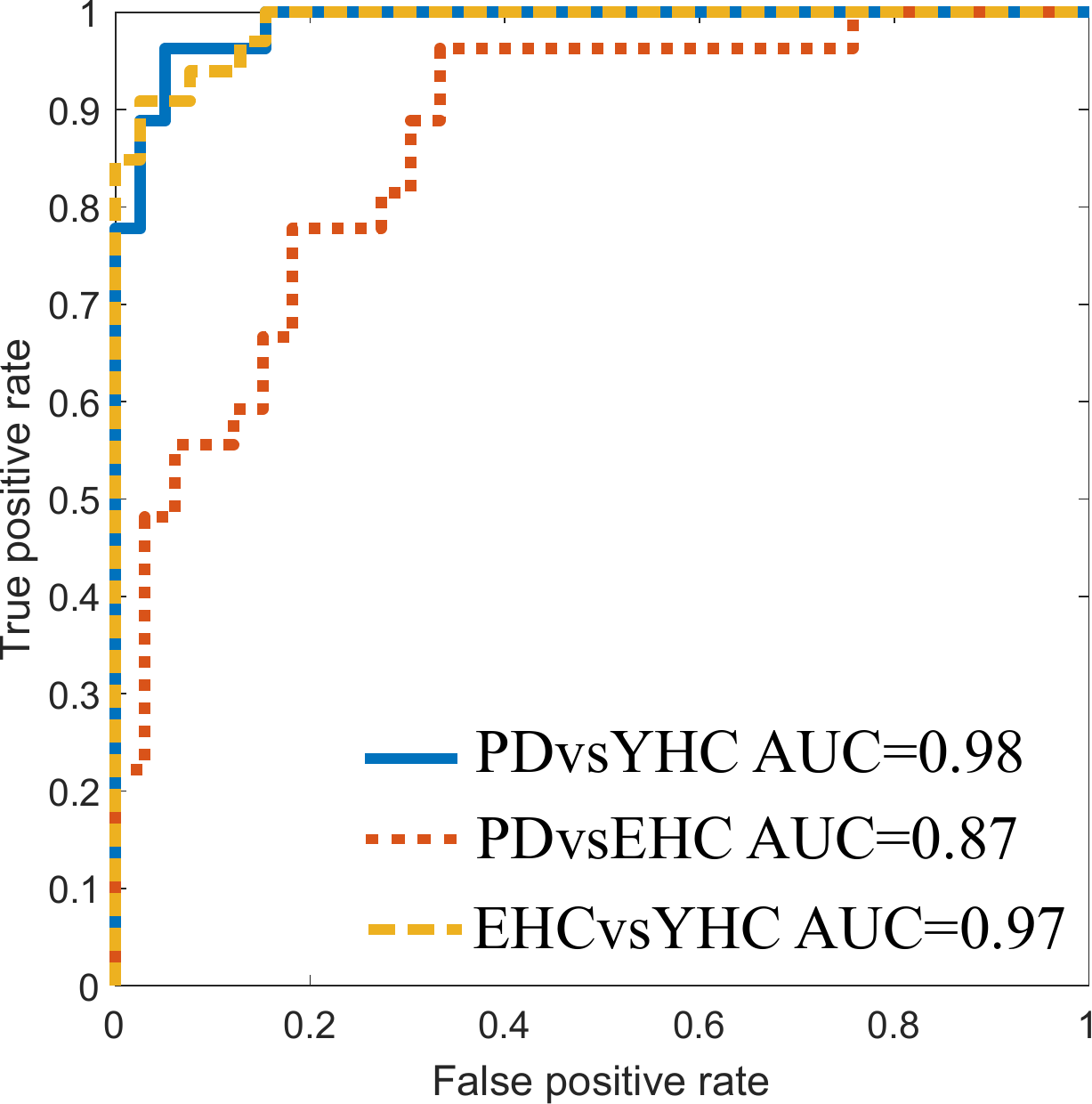}}
\subfigure[Neuromotor.]{\includegraphics[width=0.23\textwidth]{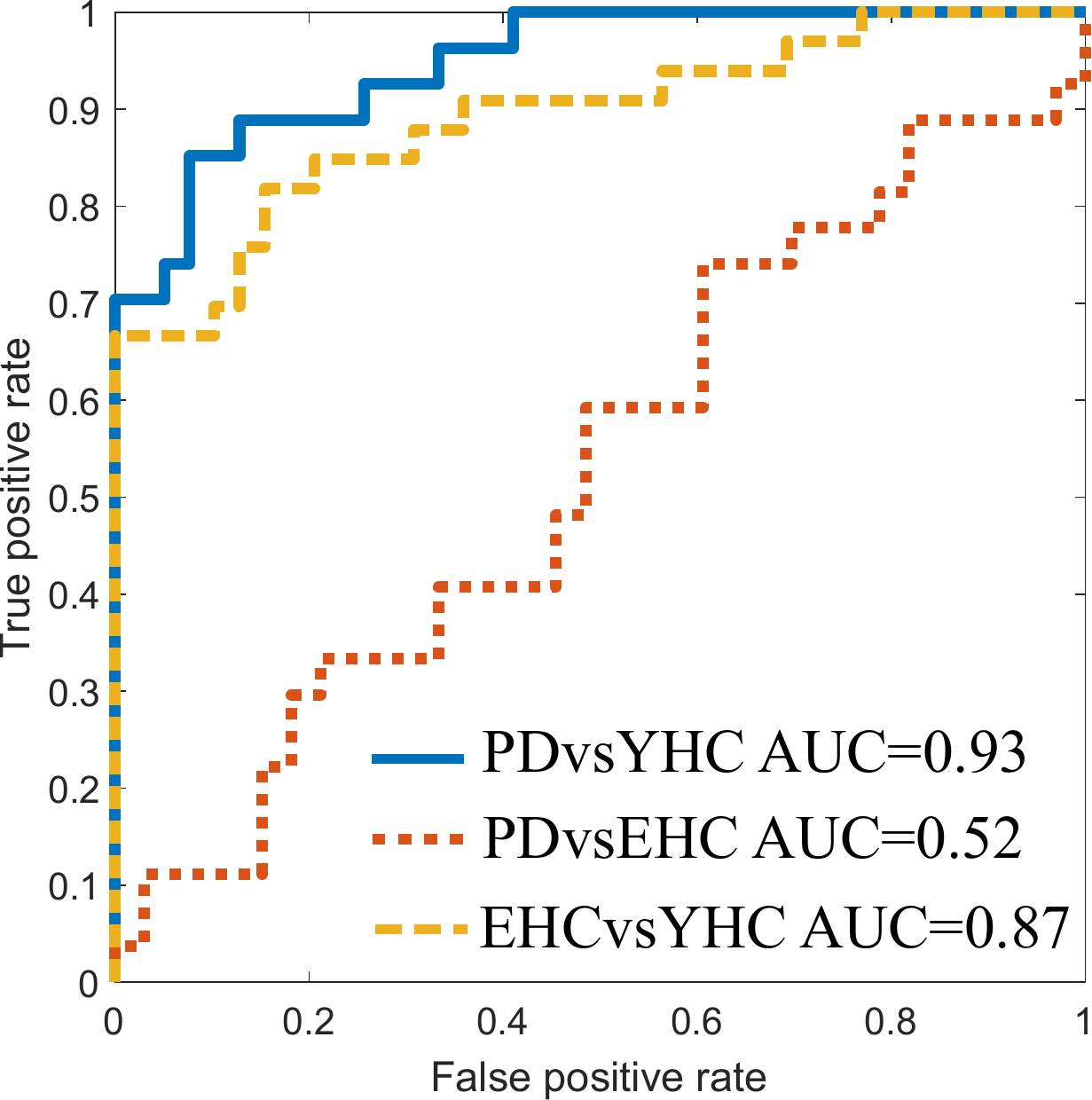}}
\subfigure[All features.]{\includegraphics[width=0.23\textwidth]{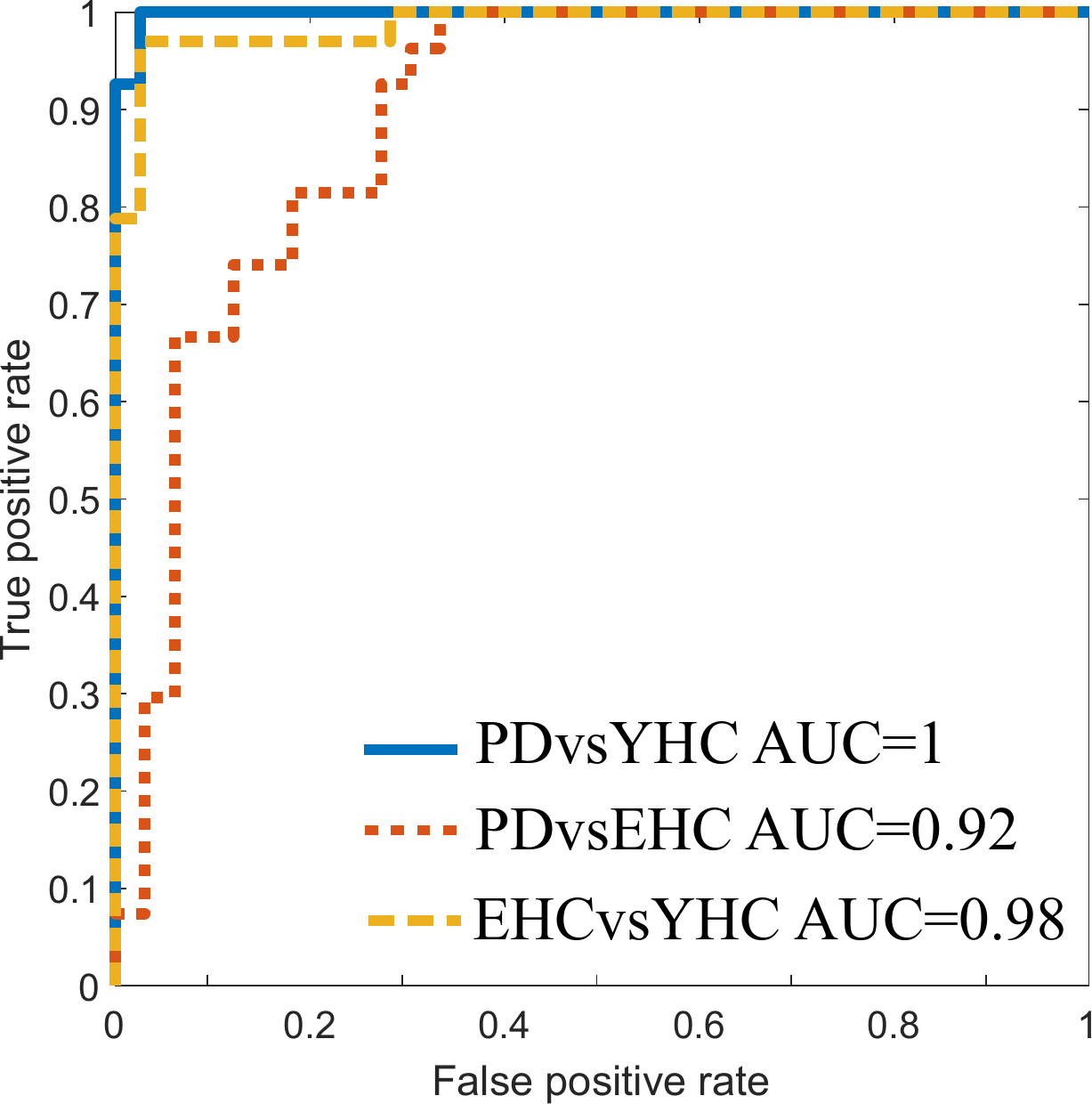}}
\caption{ROC curves YHC vs PD, EHC vs PD, and EHC vs YHC (SVM-RBF classifier). Area Under the Curve AUC=1 for perfect classification.}\label{fig:cinematicas}
\end{figure}

Figure~\ref{fig:cinematicas} shows the receiver operating characteristics curves (ROC) that result
from the analysis of the classification between PD and HC subjects. The four curves correspond
to the results obtained with the RBF-SVM considering three different feature sets (kinematics,
nonlinear dynamics, and neuromotor) and the 
combination of all of them following the late-fusion strategy. It can be observed that the experiment classifying between YHC and PD patients presents the
best results in most of the cases. When nonlinear dynamics features are used, the results of
YHC vs PD are similar to those obtained in YHC vs EHC.
The combination of all of the models presents the best results which confirms that
complimentary information can be obtained from each feature set.
Note also that the most difficult experiment is always the classification between EHC and PD patients, which confirms other works in the literature where the effect of aging in handwriting is reported.
\section{CONCLUSIONS}
In this work we have presented a large database of online handwriting for research in Parkinson Disease (PD) diagnosis. The richness of the database is not only in the number of PD patients and control subjects, but also in the quantity and diversity of the tasks performed. The techniques used both for the extraction of features and for the optimization and classification process show the potential of online handwriting as a valid biomarker for the study of PD.  We report values of 96.9\% accuracy in the classification of PD vs YHC (Young Healthy Controls), 81.7\% in the classification of PD vs EHC (Elderly Healthy Controls), and 97.2\% in the classification of EHC vs YHC.

Handwriting is a complex task that involves different dimensions: cognitive, intelligibility, visual and motor, where different muscle groups intervene, degrees of freedom, movement of the arms, turns of the wrist, extension of the fingers, etc. Within motor activity, bradykinesia, tremor, involuntary movements, and muscle stiffness are all distinctive signs of PD that can be evaluated from handwriting tasks. Such handwriting tasks can help to identify specific characteristics of the disease in early stages for the opportune diagnosis and monitoring of the disease.

The analysis of kinematic features showed an outstanding accuracy in the identification of patients with PD. Those features were also expanded including non-linear dynamics and neuromotor features. With the neuromotor features we evaluated the correlation existing between the order issued in the central nervous system and the action performed directly by the motor system. This synergistic behavior is strongly affected by the alteration of the dopaminergic system typical of PD. In a complementary way, the analysis of nonlinear dynamics measured atypical handwriting produced by chaotic behaviors at muscular level. In practice, the considered neuromotor features did not yield satisfactory results. With other processing methods we may be more successful in generating relevant information related to neuromotor actions from the handwriting signals considered.

On the other hand, motor skills deteriorate naturally with aging, and it is crucial to differentiate such degradation with respect to the motor degradation caused by PD. In this work we have also used the different tasks to study aspects related to aging by considering both young and elderly control subjects. We advocate to use both types of controls in this kind of studies to distinguish between the damage caused by the disease and age-specific effects.

As future work we propose to carry out a taxonomic study of each of the tasks based on the experience of medical specialists, to complement the proposed analyses with offline writing analysis from historical samples of patients, to expand the database periodically to perform analysis of the progression of the disease, to explore other techniques of classification and feature extraction, to include techniques of dimensionality reduction and feature selection.
\clearpage
\newpage
\addtolength{\textheight}{-20cm}

\end{document}